%% file: main.tex
\documentclass[conference,a4paper]{IEEEtran}
\IEEEoverridecommandlockouts
\IEEEoverridecommandlockouts
\IEEEpubid{\makebox[\columnwidth]
{\hfill }
\hspace{\columnsep}\makebox[\columnwidth]{ }}
\usepackage{tikz}
\usetikzlibrary{calc}
\usepackage{xcolor}
\definecolor{myred}{RGB}{255,0,0}
\usepackage{graphicx}
\usepackage{booktabs}
\usepackage{array}
\usepackage{caption}
\usepackage{amsmath,amssymb,amsfonts}
\usepackage{balance}
\usepackage{subcaption}
\usepackage{algorithmic}
\usepackage{graphicx}
\usepackage{textcomp}
\usepackage{xcolor}
\usepackage{comment}
\def\BibTeX{{\rm B\kern-.05em{\sc i\kern-.025em b}\kern-.08em
    T\kern-.1667em\lower.7ex\hbox{E}\kern-.125emX}}

\usepackage{enumitem}    
\usepackage{glossaries}
\usepackage{url}
\usepackage[normalem]{ulem}
\usepackage[colorlinks,linkcolor={black},citecolor={blue!80!black},urlcolor={blue!80!black}]{hyperref}

\loadglsentries{abbreviation}

\glsdisablehyper

\usepackage{booktabs} 
\usepackage{mathtools}
\usepackage{cite}

\DeclarePairedDelimiterX{\expectarg}[1]{[}{]}{%
  \ifnum\currentgrouptype=16 \else\begingroup\fi
  \activatebar#1
  \ifnum\currentgrouptype=16 \else\endgroup\fi
}
\usepackage{multirow}

\makeatletter
\let\old@ps@headings\ps@headings
\let\old@ps@IEEEtitlepagestyle\ps@IEEEtitlepagestyle
\def\confheader#1{%
\def\ps@headings{%
\old@ps@headings%
\def\@oddhead{\strut\hfill#1\hfill\strut}%
\def\@evenhead{\strut\hfill#1\hfill\strut}%
}%

\def\ps@IEEEtitlepagestyle{%
\old@ps@IEEEtitlepagestyle%
\def\@oddhead{\strut\hfill#1\hfill\strut}%
\def\@evenhead{\strut\hfill#1\hfill\strut}%
}%
\ps@headings%
}
\makeatother

\confheader{%
}
\begin{document}

\title{Error Mitigation for TDoA UWB Indoor Localization using Unsupervised Machine Learning 
}
\author{\IEEEauthorblockN{Phuong Bich Duong\IEEEauthorrefmark{1} \IEEEauthorrefmark{2},
Ben Van Herbruggen\IEEEauthorrefmark{1}, 
Arne Broering \IEEEauthorrefmark{2} ,
Adnan Shahid \IEEEauthorrefmark{1} , 
Eli De Poorter \IEEEauthorrefmark{1}
}

\IEEEauthorblockA{\IEEEauthorrefmark{1} IDLab, Department of Information Technology, Ghent University, Ghent, Belgium}
  \IEEEauthorblockA{\IEEEauthorrefmark{2} Siemens AG, Munich, Germany}
}

\maketitle

\begin{tikzpicture}[remember picture,overlay]
    \node [draw, fill=white, text width= 500] at ($(current page.south)+(0,1.5cm)$) {
 This work has been submitted to the IEEE for possible publication. Copyright may be transferred without notice, after which this version may no longer be accessible.
    };
\end{tikzpicture}
 
\begin{abstract} 
Indoor positioning systems based on \gls{UWB} technology are gaining recognition for their ability to provide cm-level localization accuracy. However, these systems often encounter challenges caused by dense multi-path fading, leading to positioning errors. To address this issue, in this letter, we propose a novel methodology for unsupervised anchor node selection using \gls{DEC}. Our approach uses an \gls{AE} before clustering, thereby better separating \gls{UWB} features into separable clusters of UWB input signals. We furthermore investigate how to rank these clusters based on their cluster quality, allowing us to remove untrustworthy signals. Experimental results show the efficiency of our proposed method, demonstrating a significant 23.1\% reduction in \gls{MAE} compared to without anchor exclusion. Especially in the dense multi-path area, our algorithm achieves even more significant enhancements, reducing the MAE by 26.6\% and the 95th percentile error by 49.3\% compared to without anchor exclusion.
\end{abstract}


\section{Introduction}
\gls{UWB} technology has gained popularity in indoor positioning due to its ability to provide high resolution and accuracy. This is achieved through the use of a large transmission bandwidth, which offers resistance against multipath fading and enables high temporal resolution with low transmission power. Various localization techniques, such as \gls{TWR}, \gls{TDoA} leverage the precise timing resolution of \gls{UWB} to accurately calculate positions. \gls{TWR} techniques measure distance by measuring the time of flight between two nodes, making them robust against clock drift errors and eliminating the need for synchronization between anchors \cite{jiang2007asymmetric}. \gls{TDoA} techniques, on the other hand, calculate position based on the time difference of arrival between different anchor pairs using a single \gls{UWB} transmission from the tag \cite{choi2018uwb}. However, \gls{NLoS} conditions in real-world scenarios introduce errors in measured data and significantly degrade positioning accuracy. To address this issue, it is crucial to assess the channel state and minimize the effect of \gls{NLoS} conditions, as relying solely on distance measurements in \gls{NLoS} environments is unreliable. 
There have been many ongoing studies to reduce the position estimation error, especially in harsh multipath environments, mainly focusing on two approaches: non-data-driven and data-driven. The authors in \cite{chen2022channel} propose a non-data-driven approach to evaluate the quality of \gls{CIR} based on \gls{CIR} characteristics, then later choose only good quality \gls{CIR}s for position calculation. While the approach achieves high accuracy, it requires labeling of the template CIR and considerable effort in determining the optimal set of empirical weights tailored to specific environmental conditions.  Recently, data-driven approach-based \gls{ML} has been applied widely in addressing \gls{NLoS} identification and error mitigation to improve the performance of indoor \gls{UWB} system. Table~\ref{table:related_work} provides an overview of several related papers applying \gls{ML} for \gls{NLoS} detection and error mitigation. The authors in \cite{jiang2020uwb,sang2020identification, liu2022uwb} apply supervised \gls{ML} to identify the channel condition either  \gls{LoS} or \gls{NLoS}. Others in \cite{van2021anchor} use supervised \gls{ML} for error correction and detection. These methods show high accuracy, yet require extensive time and effort to label the data, which can be significant drawback. To overcome the data labeling effort, the authors in \cite{fan2019non, kirmaz2021nlos} propose unsupervised machine learning methods using either \gls{GMM} or k-means clustering algorithms for \gls{NLoS} identification. They select specific features derived from \gls{CIR} as input. However, their manual selection of certain features introduces bias, and the feature extraction does not consider the full \gls{CIR} information. As shown in Table~\ref{table:related_work}, most UWB error mitigation works focus on supervised error correction and \gls{LoS}/\gls{NLoS} classification for \gls{TWR} localization approaches. This inherently limits the practical applicability of these works since (i) most commercial UWB localization systems utilize \gls{TDoA} due to its low energy consumption for the tag and (ii) it is commercially infeasible to collect labeled CIR data for each new deployment.  Our objective is not only to address these limitations by leveraging the full CIR information to cluster input data into multiple clusters, rather than limiting the clusters to only two or three as done by previous authors, but also investigate the error mitigation for \gls{TDoA} localization approach. Our goal is to minimize position estimation error by clustering \gls{CIR}s into a large number of clusters and evaluating the quality of each cluster based on predefined criteria. Subsequently, we employ only the \gls{CIR}s from high-quality clusters as input for the positioning algorithm. Partition-based clustering algorithms, such as k-means \cite{macqueen1967some} and \gls{GMM} \cite{bishop2006pattern} are commonly used to cluster input into multiple clusters. However, they are less efficient with high dimensional data input. To address this challenge, we propose an algorithm based on \gls{DEC} algorithm \cite{xie2016unsupervised} and extend it with k-means and \gls{GMM}.

\begin{table*}[!ht]
    \caption{Comparison of the proposed approach in this letter with related works}
    \centering
    \def\arraystretch{0.8}
    \begin{tabular}{@{}p{.8cm} p{.8cm} p{2cm} p{2.8cm} p{1.8cm} p{1.5cm} p{4.8cm}@{}}
    \toprule
    \textbf{Paper} & \textbf{TDoA/ TWR} & \textbf{Supervised/ unsupervised ML}& \textbf{Problems} & \textbf{CIR input}& \textbf{Experimental/ Simulation} & \textbf{Performance evaluations}. \\
    \midrule
    \cite{jiang2020uwb} & TWR & supervised   & LoS/NLoS classification & raw \gls{CIR} & Experimental & Classification accuracy = 82.14\%. \\
    \addlinespace
    \cite{sang2020identification} & TWR & supervised  & LoS/NLoS/multi-path classification & derived features & Experimental  & Classification accuracy = 91.9\%. \\
    \addlinespace
    \cite{van2021anchor} & \gls{TDoA} & supervised  & error correction & raw \gls{CIR} & Experimental  & improve \gls{MAE} 38\% in \gls{LoS} scenario and 29\% in \gls{NLoS}. \\
    \addlinespace
    \cite{liu2022uwb} & TWR & supervised  & LoS/NLoS classification & raw \gls{CIR} & Experimental  & Classification accuracy in range [90\%-97\%] corresponding to different datasets. \\
    \addlinespace            
    \cite{fan2019non} & TWR & unsupervised  & LoS/NLoS classification & derived features & Simulation  & Classification accuracy = 86.5\%.  \\
    \addlinespace
    \cite{kirmaz2021nlos}& TWR & unsupervised & LoS/NLoS classification & derived features & Experimental & Classification accuracy  in range [85\%-95\%] corresponding to different datasets. \\
    \addlinespace    
    \textbf{Our work} & \textbf{TDoA} & \textbf{unsupervised} & \textbf{error mitigation} & \textbf{raw CIR} & \textbf{Experimental} &  \textbf{improve \gls{MAE} 26.6\%} \textbf{in dense multi-path area}. \\
    \bottomrule
    \end{tabular}
    \label{table:related_work}
\end{table*}
\vspace{0.1cm}
The main contributions of this letter are as follows:
\begin{enumerate}
\setlength\itemsep{0em}
\item We are the first to propose an unsupervised machine learning framework for \gls{CIR}-based \gls{UWB} error mitigation for \gls{TDoA} system. This framework is applied for anchor selection. It uses the DEC algorithm to cluster the \gls{UWB} input signals into a predefined number of clusters. 
Afterwards, we evaluate the quality of the clusters, then only high quality clusters are used for position calculation. This approach not only reduces error in the tag position estimation but also eliminates the intensive data labeling effort.
\item We investigate the criteria to use when selecting the clusters to include for the position calculation. To this end, we calculate the distance between the first path index and peak path index of each \gls{CIR} and investigate two cluster selection criteria: utilizing the mean and standard deviation of this aforementioned distance per cluster.  
\item We compare our method with traditional clustering algorithms such as k-means, \gls{GMM}, \gls{AE}+k-means, \gls{AE}+GMM, and show that our method outperforms existing approaches.
\end{enumerate}

The remainder of this paper is organized as follows. Section II describes the data collection and preprocessing process. Section III presents methodology we used to cluster the signal and evaluate the cluster quality. The results and discussion of the performance of our approach are presented in Section IV. Finally, conclusions are provided in Section V.

\section{Data Collection and Preprocessing}
The data is collected in the IIoT-lab at Ghent University \cite{van2021anchor}. The lab is equipped with 23 Wi-Pos anchor nodes which combine the popular DW1000 \gls{UWB} receiver with a wireless sub-GHz backbone network for networking synchronization \cite{van2019wi}. The \gls{UWB} settings used in this research are channel 5, a bit rate of 850 kbps, a preamble with 512 symbols, and a \gls{PRF} of 64 MHz. All anchor nodes have a wired connection and transmit their data to a central server that stores the data (CIR, first path index, peak path index, receive power levels, timestamps, ...). 

The lab contains two major experimentation zones: one open space of 100 $m^2$ and one industrial warehouse environment of 150 $m^2$. Both zones are equipped with a mm-accurate ground-truth \gls{MOCAP} system. In the \gls{TDoA} system, one reference node is synchronizing its clock to all neighboring anchor nodes. As not all anchors are within reach of this reference node at the used \gls{UWB} settings, only the anchors that are single-hop synchronized are used. The data is collected with a mobile robot which permits, in combination with the MOCAP system, to repeat trajectories accurately. Different trajectories are specified and repeated with multiple reference nodes to synchronize the other anchors.  

\section{Methodology}
\begin{figure*}[htbp!]
    \centering
    \includegraphics[width=0.8\textwidth]{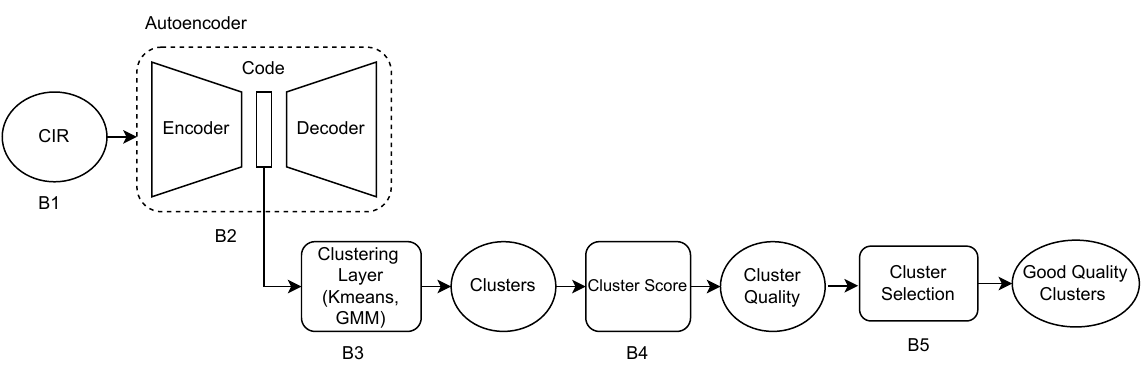}
    \caption{The architecture of the cluster selection model. The input is the raw \gls{CIR}s (B1). The \gls{AE} (B2) is used to learn the salient features of \gls{CIR} represented at code layer. Then, the code layer is used as input for  clustering layer (B3). The output of clustering layer is the clusters fed to the clustering score (B4). The clustering score outputs different cluster quality levels corresponding to the $\mu_{j}$ and $\sigma _{j}$ values of each cluster. Later, cluster selection (B5) outputs only the good quality clusters.}
    \label{fig:training_phase}
\end{figure*}

This section presents the unsupervised \gls{ML}-based \gls{DEC} algorithm we use to cluster the signals received from all anchor nodes to multiple clusters corresponding to different quality levels. Then only high quality clusters are later utilized to compute the tag position. An overall architecture of the proposed method is shown in Fig.~\ref{fig:training_phase}.

\subsection{Input Data}
The input data is  raw \gls{CIR} as shown by block B1 in the model. The CIR describes the channel propagation conditions of a radio signal and can be represented as below:
\begin{equation}\label{eq:cir}
    h(t) = \sum_{i = 1}^{\textnormal{n}} \alpha_i \delta(t - \tau_{i}) + n(t), 
\end{equation}
where $n(t)$ represents the additive white Gaussian noise, $\alpha_i$ are the coefficients of path delays denoted by $\delta(t - \tau_{i})$, and $n$ is the total number of multi-path components. 

\subsection{Deep Embedded Clustering}
The \gls{DEC} is the coupling of the \gls{AE}'s code layer with the clustering layer (as shown in Figure \ref{fig:training_phase}). The \gls{DEC} includes 2 main parts: \gls{AE} and clustering layer.
\subsubsection{Autoencoder}
The \gls{AE} as block B2 in the model is a neural network used for dimension reduction and feature learning. We use a convolutional \gls{AE} consisting of an encoder and a decoder. The encoder is a nonlinear function $\textbf{z}_{i}=f(\textbf{x}_{i})$ that maps the input vector $\textbf{x}_{i}$ into the code layer $\textbf{z}_{i}$. This code layer represents salient features of the input data. The decoder is  also considered as a nonlinear function $\bar{\textbf{x}}_{i} = g({\textbf{z}}_{i})$ that reconstructs the input vector $\bar{\textbf{x}}_{i}$ from the latent vector ${\textbf{z}}_{i}$. The learning procedure of an \gls{AE} minimizes the mean square error of reconstruction loss $ L_{r}(\textbf{x}_{i},g(f(\bar{\textbf{x}}_{i})))$. By minimizing the reconstruction loss, the encoder generates the code layer with dimensional reduction. This causes instances of the same class that have semantically similar features, to be closely grouped in the latent space. As a result, clustering within the latent space, rather than directly with raw data, demonstrates increased effectiveness and significantly decreases the computational complexity of the clustering algorithms.

\subsubsection{Clustering Layer}
The clustering layer as block B3 in the model is connected to the output of the encoder, i.e., the code layer containing the latent feature vector ${z}_{i}$. The number of $k$ clusters is defined in advance. We initialize the cluster centroids ($l _{j}$) in the feature space using either the k-means or \gls{GMM} clustering algorithm. Then we use the student’s t-distribution \cite{van2008visualizing} to calculate  the membership probabilities ($q_{ij}$) between the cluster centroids ($l _{j}$) and latent feature vector ${z}_{i}$ as follows: 
\begin{equation}\label{eq:t_sne}
q_{ij} = \frac{(1+\left \| z_{i}-l _{i}\right \|^{^{2}})^{-1}}{\sum _{j}(1+\left \| z_{i}-l _{j} \right \|^{2})^{^{-1}}}.
\end{equation}
The membership probabilities are used to compute an auxiliary target distribution $p_{ij}$ as follows:
\begin{equation}\label{eq:aux_func}
p_{ij}=\frac{q_{ij}^{2}/\sum _{j}q_{ij}}{\sum_{j}(q_{ij^{2}}/\sum_{i}q_{ij}) }.
\end{equation}

The clustering is performed by minimizing the Kullback–Leibler (KL) divergence between the soft assignments, $q_{ij}$, and the target distribution $p_{ij}$. The clustering loss, $L_{c}$ is defined as:
\begin{equation}\label{eq:KL}
L_{c}=KL(P\left |  \right |Q)=\sum_{i}\sum_{j} p_{ij} log(\frac{p_{ij}}{q_{ij}}).
\end{equation}

\subsubsection{Training Procedure}
The entire network model is trained by minimizing the loss function that is a weighted combination of the \gls{AE} loss ($L_{r}$) and clustering loss ($L_{c}$), as follows:
\begin{equation}\label{eq:loss_model}
L_{m}=(1-\lambda )L_{r}+\lambda L_{c},
\end{equation}
where $\lambda \in [0,1]$ is the the weight of the loss function. 
The training procedure includes 2 steps:
\begin{itemize}
\setlength\itemsep{0em}
\item Step 1: The \gls{AE} is trained only with the reconstruction loss by setting $\lambda =0$. This focuses on accurately reconstructing the input data and learning salient features at the code layer.
\item Step 2: The \gls{AE} and clustering layer are fine-tuned. The encoder extracts low-dimensional features, and cluster centroids are initialized using k-means and \gls{GMM}. Through the fine-tuning process, the cluster centroids are adjusted, and the \gls{AE}'s weights are updated. This process facilitates the learning of clustering-friendly features, leading to improved clustering effectiveness.
\end{itemize} 

\subsection{Determining Cluster Quality Scores}
\subsubsection{Determining Cluster Quality}
The quality of clusters is done by block B4, cluster score. The quality of the \gls{CIR} is crucial for accurate tag position estimation. Depending on the channel conditions, different quality levels of a \gls{CIR} can be observed. To evaluate the quality of cluster, the distance between the position of the detected first path $pp_{idx}$ and peak path $fp_{idx}$ is considered. The distance $d_{i}$ can be calculated using the following formula:
\begin{equation}\label{eq:distance}
        d_{i}= (pp_{idx_i}-fp_{idx_i}) * t_{s_i} * c,
\end{equation}
 where $t_{s_i}$  and $c$ are \gls{CIR} time and speed of light respectively.
 Once cluster labels are received from the clustering layer, the mean $\mu_{j}$ and standard deviation $\sigma _{j}$ of the distance $d_{ji}$ of all \gls{CIR} samples in each cluster label $k_{j}$  are calculated.
We utilize the values of  $\mu_{j}$ and $\sigma _{j}$ as the criteria to assess the quality of the clusters. Generally, higher values of  $\mu_{j}$ and $\sigma _{j}$ indicate lower cluster quality. The standard deviation $\sigma _{j}$ provides information about the dispersion of data around the mean $\mu_{j}$. Therefore, higher $\sigma _{j}$ values indicate more scattered \gls{CIR}s within the cluster. We assign scores to the clusters based on their $\mu_{j}$ and $\sigma _{j}$ values, where lower scores correspond to higher quality and higher scores indicate lower quality.

\subsubsection{Cluster Selection}
The selection of good quality clusters are done by cluster selection as block B5 in model. The output of cluster score, cluster quality is scored from lowest to highest values corresponding to their $\mu_{j}$ and $\sigma _{j}$ values. The clusters with lower scores indicate better quality which are subsequently selected to be fed to a positioning algorithm.
 
\subsection {Evaluation of position estimate}
To evaluate the performance of the clustering method, we derive the positioning
error of the tag position estimation. We first apply deep clustering for the whole dataset, then calculate the $\mu _{j}$ and  $\sigma _{j}$ for each cluster. Afterward, we sort the clusters from smallest to highest $\mu _{j}$ and $\sigma _{j}$, and score the clusters from $0$ to $k$ values accordingly. To predict the position of the tag at each time index, we exclude the signal that has the high score and consider only the signals from anchor nodes that have low scores as the input for the \gls{TDoA} positioning algorithm.

\section{Evaluation}
In this section, we evaluate the performance of error mitigation by using our proposed approach. We derive the positioning error to measure the performance of the algorithms. We use the \gls{TDoA} least squares algorithm that considers the \gls{TDoA} of high quality signals and determines the positioning error (as the Euclidean distance) against the ground-truth of the measurements. We also compare the results of our approach with traditional clustering algorithms k-means, \gls{GMM}, \gls{AE}+k-means and \gls{AE}+ \gls{GMM}.
\subsection{Clustering Performance Visualization and Cluster Score}
\begin{figure}[!t]
    \centering
    \begin{subfigure}[b]{0.22\textwidth}
        \centering
        \includegraphics[width=\textwidth]{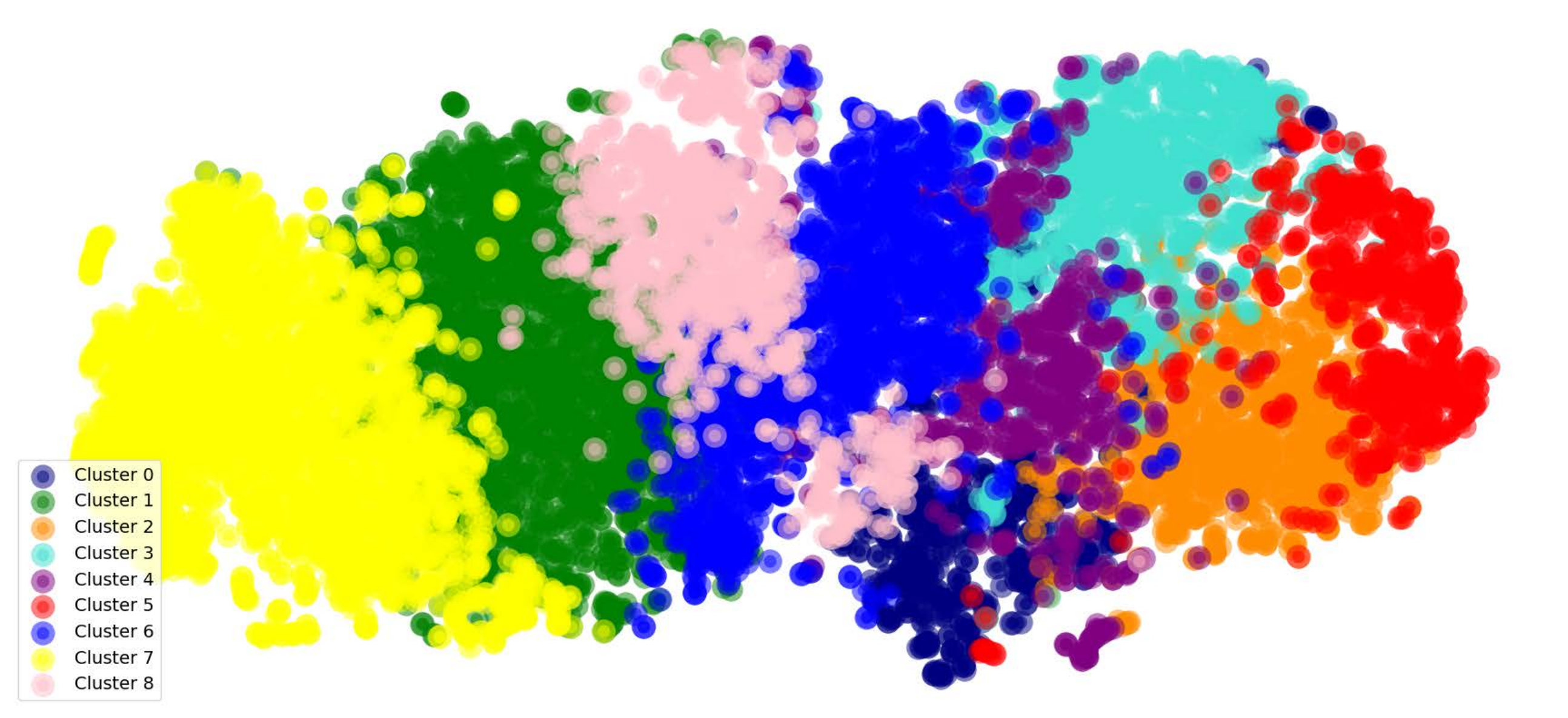}
        \caption{}
        \label{fig:k_means}
    \end{subfigure}
    \hfill    
    \begin{subfigure}[b]{0.22\textwidth}
        \centering
        \includegraphics[width=\textwidth]{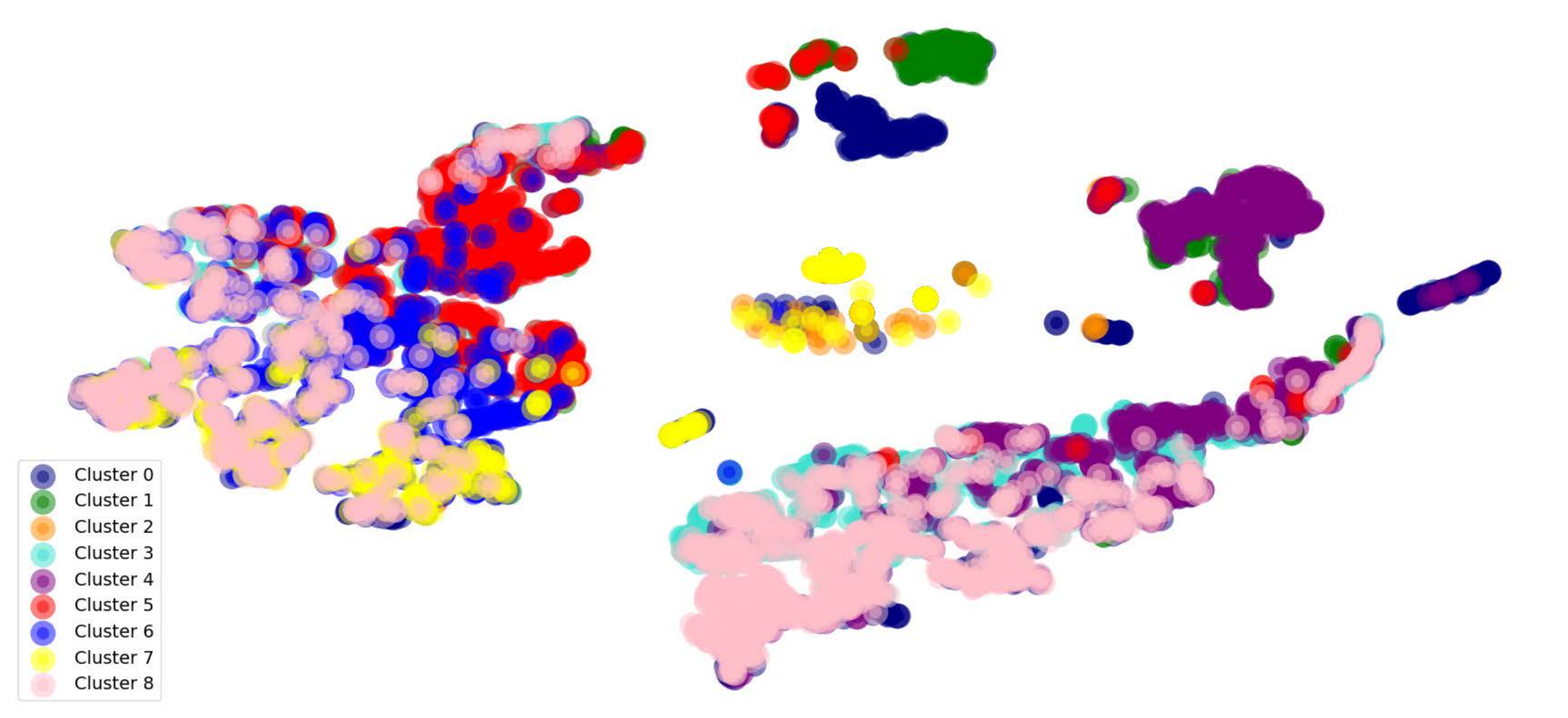}
        \caption{}
        \label{fig:ae+kmeans}
    \end{subfigure}
    \hfill
      
    \begin{subfigure}[b]{0.22\textwidth}
        \centering
        \includegraphics[width=\textwidth]{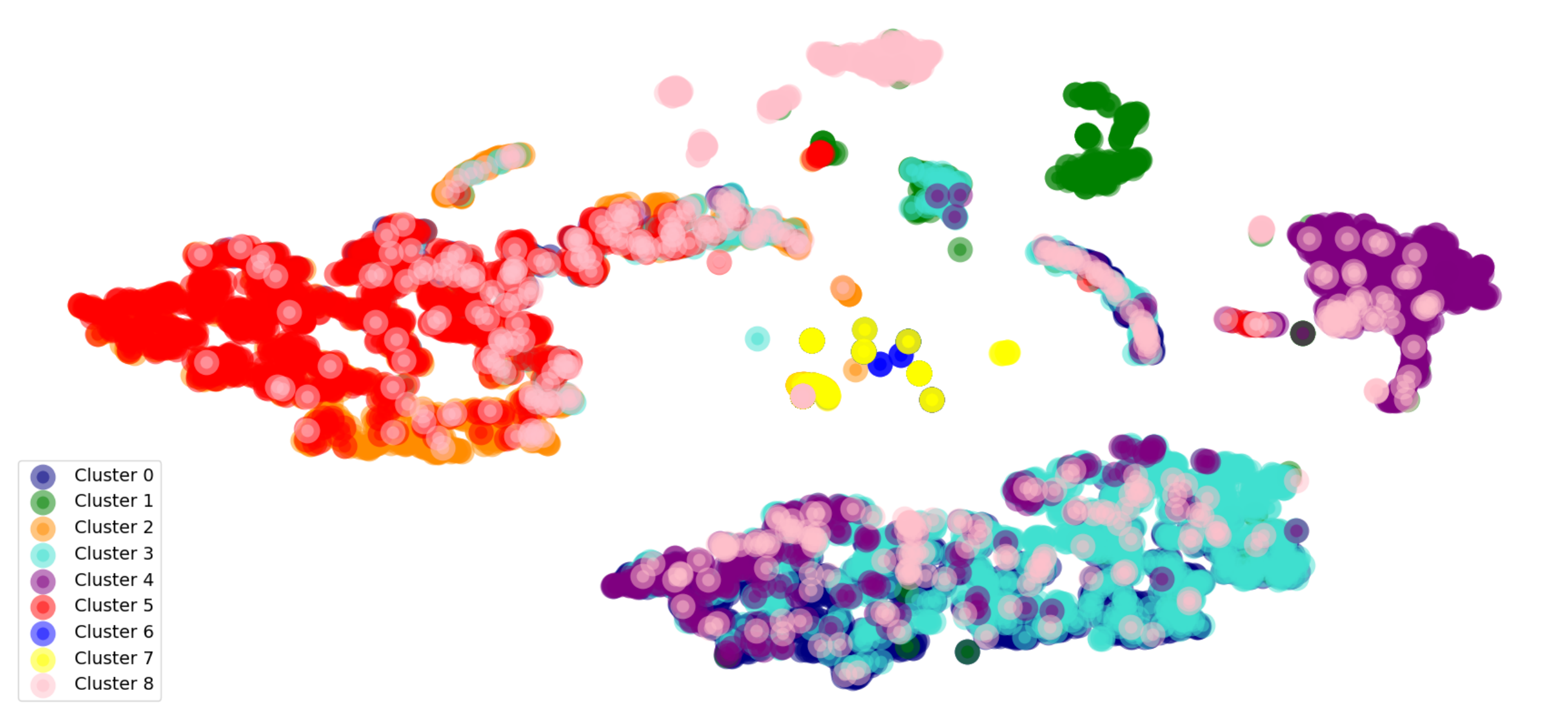}
        \caption{}
        \label{fig:dec+gmm_combi_training_iter=1500}
    \end{subfigure}
    \hfill
    \begin{subfigure}[b]{0.22\textwidth}
        \centering
        \includegraphics[width=\textwidth]{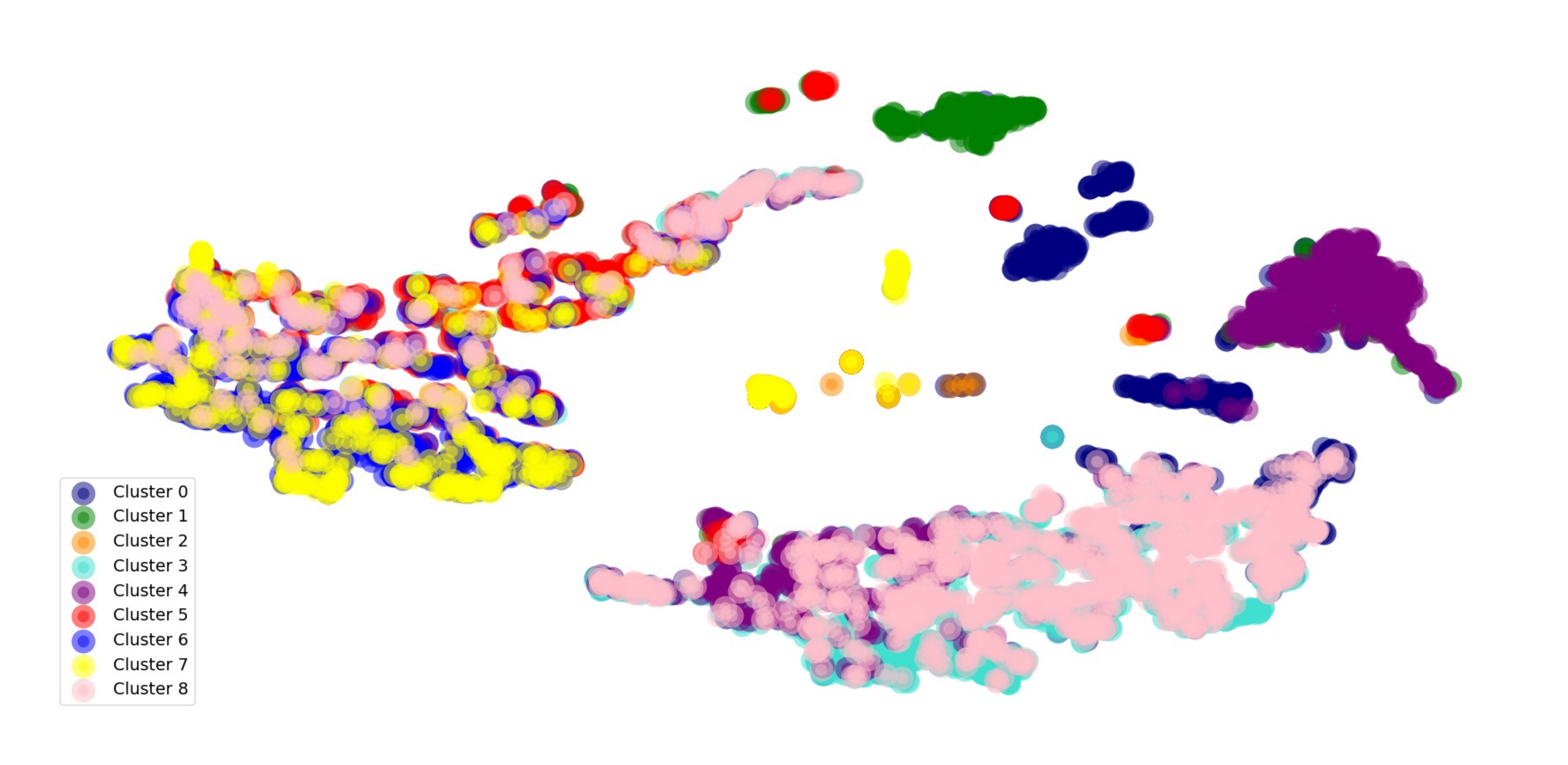}
        \caption{}
        \label{fig:dec+kmeans_combi_training_iter=1500}
    \end{subfigure}

    \caption { \gls{CIR} input data clustered to 9 clusters using (a) k-means, (b) AE+k-means, (c) DEC+GMM with iteration = 1500, (d) DEC+k-means with iteration = 1500 }
    \label{fig:dec}
\end{figure}

 The data is collected from $23$ anchor nodes located in the experimentation zones. However, at each time index , not all anchor nodes receive signal originated from tag. In fact, a maximum of $9$ signals were recorded from synchronized anchor nodes. Hence we choose the predefined number of cluster, $k=9$, for our analysis. Fig.~\ref{fig:dec} illustrates the performance of different clustering approaches\footnote{To reduce the data to two dimensions for visualization, we used the \gls{T-SNE} method \cite{van2008visualizing}.}.
Fig.~\ref{fig:k_means} shows k-means clustering on the raw data. After the pretraining of the \gls{AE}, the \gls{CIR} raw data was transformed into a feature space with a lower dimension and afterward k-means clustering was applied as shown in Fig.~\ref{fig:ae+kmeans}. It slightly increases the pairwise distances between cluster centroids. The low quality clusters according to the $\sigma _{j}$ value criterion are slightly separated from the other clusters. Next, we perform the fine-tuning and update the \gls{AE} weights every 100 iterations. This gradually increases the distance from the low quality clusters to others as more useful features are learned and optimized for the clustering task. Fig.~\ref{fig:dec+gmm_combi_training_iter=1500} and ~\ref{fig:dec+kmeans_combi_training_iter=1500} visualize the \gls{GMM} and k-means embedded clustering at 1500 iterations. Fig.~\ref{fig:dec+kmeans_combi_training_iter=1500} shows the low quality clusters including cluster \textit{1,0,4} corresponding to the $\sigma _{j}$ value score that are well separated from the others. The \gls{GMM}-based embedded clustering has worse performance compared to the k-means-based version, since there are still mixed data points from other low quality clusters (see cluster \textit{1,0,4}).

\subsection{Position estimate}
\subsubsection{Tag position estimation over the whole trajectory}
Table~\ref{table:pos_accuracy} presents the positioning estimation results for each clustering method corresponding to $\mu _{j}$ and $\sigma _{j}$ selection criteria, wherein one signal with the highest score value is systematically excluded at each time index for position calculation. With the selection criterion $\sigma _{j}$, the traditional method of clustering using \gls{GMM} fails to mitigate positioning errors, whereas k-means demonstrates a better performance by reducing both the \gls{MAE} from 0.26m to 0.21m and the 95 percentile error from 0.93m to 0.66m. The integration of the code layer from the \gls{AE} with clustering algorithms k-means and \gls{GMM} shows significant gains, especially when considering the reporting of 95th percentile of the error. These combined approaches are able to significantly diminish high-error position estimations. When initializing the cluster centroids of \gls{DEC} using the k-means and \gls{GMM} clustering algorithms and fine-tuning the \gls{AE} and clustering layer over the iterations,  \gls{DEC}+k-means effectively separates clusters with high $\mu _{j}$ and $\sigma _{j}$ from the other clusters compared to \gls{DEC}+\gls{GMM} (see Fig .~\ref{fig:dec+kmeans_combi_training_iter=1500} and Fig .~\ref{fig:dec+gmm_combi_training_iter=1500}). Hence, a reduction in \gls{MAE} is achieved, reaching values of 0.20m and 0.21m, respectively. Thus, these methods result in a substantial decrease in 95th percentile of the error, lowering them from 0.93m to 0.51m and 0.52m for k-means and \gls{GMM}, respectively. The DEC+k-means combination demonstrates superior performance, as it not only decreases \gls{MAE} from 0.26m to 0.20m but also achieves a significant 45\% reduction in 95th percentile of error, decreasing the tag position estimation error from 0.93m to 0.51m. The table results also indicate that we achieve better performance when using $\sigma _{j}$ as the score criterion compared to $\mu _{j}$ . 

\begin{figure}[!t]
    \centering
    \includegraphics[width=0.45\textwidth]{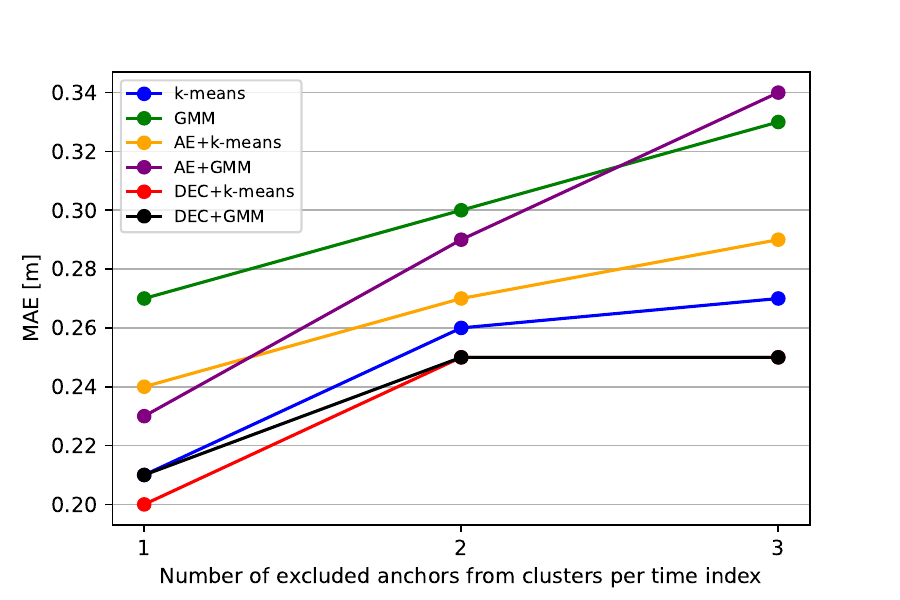}
    \caption{Impact of excluding additional number of anchors from clusters on \gls{MAE}.}
    \label{fig:MAE}
\end{figure}

We also investigate the impact of excluding additional number of anchor from low quality clusters on \gls{MAE}. Fig.~\ref{fig:MAE} shows that higher numbers of signal exclusion results in \gls{MAE} increase. Hence, we achieve the best result when excluding only the highest score signal for all methods. 
\begin{table}[!ht]
    \caption{Positioning accuracy (in m) with $\sigma _{j}$ and $\mu _{j}$ as selection criteria}
    \centering
    \def\arraystretch{0.9}
    \begin{tabular}{p{3cm} *{4}{>{\centering\arraybackslash}p{0.9cm}}}
        \toprule
        \textbf{Method} & \multicolumn{2}{c}{$\sigma _{j}$} & \multicolumn{2}{c}{$\mu _{j}$} \\
        \cmidrule(lr){2-3} \cmidrule(lr){4-5}
        & \textbf{MAE} & \textbf{95th} & \textbf{MAE} & \textbf{95th} \\
        \midrule
        Without anchor exclusion & 0.26  & 0.93 & 0.26  & 0.93 \\
        k-means & 0.21  & 0.66 & 0.44  & 2.16 \\
        GMM & 0.27 & 1.05 & 0.26 &  0.94 \\
        AE+k-means & 0.22 &  0.55 & 0.24  & 0.75 \\
        AE+GMM & 0.23 & 0.73 & 0.23 & 0.73 \\
        DEC+k-means & \textbf{0.20} & \textbf{0.51} & 0.21 & 0.52 \\
        DEC+GMM & 0.21 & 0.52 & 0.21  & 0.52 \\
        \bottomrule
    \end{tabular}
    \label{table:pos_accuracy}
\end{table}

\subsubsection{Tag position estimation at dense multi-path area}
\begin{table}[!ht]
	\caption{ Positioning accuracy (in m) in dense multi path area with the $\sigma _{j}$ as criterion
	}
	
	\begin{center}
			\def\arraystretch{1.0}
			\begin{tabular}{ p{3cm} p{0.9cm}  p{0.90cm}  p{0.90cm}  p{0.9cm}  }
				\hline
				\textbf{Method}&\textbf{MAE} &  \textbf{75th} &  \textbf{90th} & \textbf{95th}  \\ 
                \hline
                Without anchor exclusion    & 0.45 & 0.46 & 1.04  & 1.47 \\
                k-means     & 0.38  & 0.41 & 0.71 & 1.01 \\               
                \gls{GMM}     & 0.48 & 0.52 & 1.11 & 1.54 \\
                AE+k-means     & 0.39 & 0.39  & 0.76 & 1.05 \\
                AE+\gls{GMM}     & 0.37  & 0.39  & 0.74 & 1.05 \\
                \gls{DEC}+k-means     & \textbf{0.33} & \textbf{0.34}  & \textbf{0.55} & \textbf{0.74} \\
                \gls{DEC}+\gls{GMM}    &  0.35  & 0.35   & 0.6 & 0.81 \\
                \hline
			\end{tabular}\label{table:dense_multipath}
	\end{center}
\end{table}
To assess the effectiveness of our methodology, we evaluate the dense multipath area, i.e., the industrial warehouse environment where signal distortion and \gls{NLoS} signals are dominant. Table~\ref{table:dense_multipath} shows the tag position estimation in this area. The results show that the \gls{DEC}+k-means method also yields superior tag position estimation with lowest \gls{MAE} and 95th percentile compared to other methods.

\section{Conclusion}
In this letter, we proposed an unsupervised machine learning using \gls{DEC} for anchor selection to reduce the error for \gls{TDoA} \gls{UWB} systems. Our proposed approach successfully reduces 23,1\% of the \gls{MAE} and 45,2\% of 95th percentile error for whole tag trajectory and reduces 26,6\% \gls{MAE} and 49,3\% of 95th percentile error in the dense multi-path area compared to without anchor exclusion. Importantly, our approach eliminates the need for intensive data labeling efforts, ensuring high efficiency and practicality. Additionally, our approach is applicable at the anchor level, making it suitable for distributed indoor positioning systems.

\section*{Acknowledgment}
This work was funded by the German Federal Ministry of Education and Research (BMBF) project 6G-ANNA (16KISK098).

\bibliographystyle{IEEEtran}
\bibliography{bibliography}

\end{document}

%% file: abbreviation.tex
\newacronym{ADMA}{ADMA}{Automotive Dynamic Motion Analyzer}
\newacronym{AE}{AE}{Auto Encoder}
\newacronym{AoA}{AoA}{Angle of Arrival}

\newacronym{BS}{BS}{base station}
\newacronym{BLE}{BLE}{Bluetooth Low Energy}

\newacronym{CIR}{CIR}{channel impulse response}
\newacronym{C/N0}{$C/N_0$}{carrier-to-noise density ratio}
\newacronym{CDF}{CDF}{cumulative distribution function}

\newacronym{DL}{DL}{deep learning}
\newacronym{DOP}{DOP}{Dilution of Precision}
\newacronym{DC}{DC}{Deep clustering}
\newacronym{DEC}{DEC}{deep embedded clustering}
\newacronym{DDoA}{DDoA}{distance difference of arrival}



\newacronym{GMM}{GMM}{Gaussian mixture model }

\newacronym{IMU}{IMU}{Inertial Measurement Unit}
\newacronym{IoT}{IoT}{Internet of Things}
\newacronym{IDEC}{IDEC}{Improved Deep Embedding Clustering}

\newacronym{KF}{KF}{Kalman filter}

\newacronym{LoS}{LoS}{line of sight}
\newacronym{LTE}{LTE}{Long-Term Evolution}

\newacronym{ML}{ML}{machine learning}
\newacronym{MAE}{MAE}{ mean absolute error}
\newacronym{MOCAP}{MOCAP}{motion capture}


\newacronym{NLoS}{NLoS}{non-line of sight}



\newacronym{PDF}{PDF}{probability density function}
\newacronym{PRF}{PRF}{pulse repetition frequency}


\newacronym{RMS}{RMS}{Root-mean square}

\newacronym{RF}{RF}{radio frequency}
\newacronym{pdf}{pdf}{probability distribution function}
\newacronym{RMSE}{RMSE}{root mean squared error}

\newacronym{SNR}{SNR}{Signal-to-noise ratio}
\newacronym{SW}{SW}{software}

\newacronym{TDoA}{TDoA}{Time Difference of  Arrival}
\newacronym{ToA}{ToA}{time of arrival}
\newacronym{T-SNE}{T-SNE}{T-distributed stochastic neighbor embedding}
\newacronym{TWR}{TWR}{Two-Way Ranging}

\newacronym{UWB}{UWB}{Ultra-wideband}